\documentclass{article}



\usepackage{arxiv}

\usepackage[utf8]{inputenc} 
\usepackage[T1]{fontenc}    
\usepackage{hyperref}       
\usepackage{url}            
\usepackage{booktabs}       
\usepackage{amsfonts}       
\usepackage{nicefrac}       
\usepackage{microtype}      
\usepackage{lipsum}
\usepackage{graphicx}
\usepackage{algorithm}
\usepackage[noend]{algpseudocode}
\usepackage{amsmath,amssymb,latexsym,float,epsfig}
\usepackage{mathtools,breqn,amsmath}
\usepackage{cite}
\usepackage{caption}
\usepackage{subcaption}
\usepackage{graphicx}
\usepackage{booktabs}
\usepackage{svg}
\usepackage{color}
\usepackage{tabularray}

\graphicspath{ {./images/} }

\hypersetup{
urlcolor=magenta
}
\title{An Analysis of Kalman Filter based Object Tracking Methods for Fast-Moving Tiny Objects}

\author{
  Prithvi Raj Singh\textsuperscript{\thanks{Corresponding Author} 1,3},
  Raju Gottumukkala\textsuperscript{2,3},
  Anthony Maida\textsuperscript{2,4}\\
  \textsuperscript{1}McNeese State University, \texttt{psingh8@mcneese.edu} \\
  \textsuperscript{2}University of Louisiana at Lafayette, \texttt{raju@louisiana.edu}, \texttt{maida@louisiana.edu} \\
  \textsuperscript{3}Informatics Research Institute (IRI), \textsuperscript{4}Center for Advanced Computer Studies (CACS) \\
}

\begin{document}
\maketitle

\begin{abstract}

Unpredictable movement patterns and small visual mark make precise tracking of fast-moving tiny objects like a racquetball one of the challenging problems in computer vision. This challenge is particularly relevant for sport robotics applications, where lightweight and accurate tracking systems can improve robot perception and planning capabilities. While Kalman filter-based tracking methods have shown success in general object tracking scenarios, their performance degrades substantially when dealing with rapidly moving objects that exhibit irregular bouncing behavior. In this study, we evaluate the performance of five state-of-the-art Kalman filter-based tracking methods—OCSORT, DeepOCSORT, ByteTrack, BoTSORT, and StrongSORT—using a custom dataset containing 10,000 annotated racquetball frames captured at 720p-1280p resolution. We focus our analysis on two critical performance factors: inference speed and update frequency per image, examining how these parameters affect tracking accuracy and reliability for fast-moving tiny objects. Our experimental evaluation across four distinct scenarios reveals that DeepOCSORT achieves the lowest tracking error with an average ADE of 31.15 pixels compared to ByteTrack's 114.3 pixels, while ByteTrack demonstrates the fastest processing at 26.6ms average inference time versus DeepOCSORT's 26.8ms. However, our results show that all Kalman filter-based trackers exhibit significant tracking drift with spatial errors ranging from 3-11cm (ADE values: 31-114 pixels), indicating fundamental limitations in handling the unpredictable motion patterns of fast-moving tiny objects like racquetballs. Our analysis demonstrates that current tracking approaches require substantial improvements, with error rates 3-4x higher than standard object tracking benchmarks, highlighting the need for specialized methodologies for fast-moving tiny object tracking applications.

Project Code is avilable at \url{https://github.com/Prithviraj97/KFTrackingEvaluation}
\end{abstract}


\section{Introduction}
Recent advances in object tracking have led to significant improvements in both single-object tracking (SOT) and multi-object tracking (MOT) applications, with modern systems achieving less than 5\% error rates on standard benchmarks involving larger objects. Modern tracking systems can effectively capture deep visual features of target objects, resulting in enhanced tracking performance across various scenarios. However, most of the progress around object tracking has been centered around tracking relatively large and stationary objects that follow predictable, linear motion patterns.

Tracking rapidly moving small objects, such as racquetballs or tennis balls, is more challenging problem. Racquetball can exceed speed of 150 mph and can have multiple direction changes per second creating hard to predict non-linear trajectory pattern. 
The difficulty arises from several factors: the small size of objects leads to limited visual information for feature extraction, high-speed motion causes motion blur and rapid scale changes, background clutter can easily overwhelm small targets, and frequent occlusions make continuous tracking difficult. In the specific context of racquetball, these challenges are amplified because the ball exhibits multiple short-distance bounces and ricochets off sidewalls, creating highly erratic and non-linear motion patterns with sudden acceleration changes that are difficult to predict and track reliably over extended period.

Object tracking approaches generally fall into two main categories: tracking-by-detection and tracking-without-detection. In tracking-by-detection methods, the primary focus is on robust object detection in each frame, followed by association and tracking of detected objects across all video frames. This approach typically combines deep learning-based object detection algorithms—such as Fast-RCNN ~\cite{fastrcnn}, Faster-RCNN ~\cite{fasterrcnn}, YOLO models ~\cite{wang2023yolov7, yaseen2024yolov8indepthexplorationinternal, khanam2024yolov5deeplookinternal}, or SSD ~\cite{ssd} — to locate target objects, with tracking algorithms deployed on top of the detection system to associate objects across frames while maintaining object identities and estimating future positions. 

In tracking-without-detection method, the tracking algorithm is solely responsible for tracking the target without performing detection at each frame. The algorithm itself estimates the future position, velocity, and other parameters of the target based on the given initial position and subsequently previously available information. Motion or object appearance based cues are used to track the target object. Optical flow based method~\cite{opticalflow}, mean-shift tracking~\cite{meanshift}, and correlation filters~\cite{correlationfilter} are some of the common algorithm for tracking-without-detection. However, these methods face particular challenges with tiny objects where there is limited appreance cues for feature extraction.

For tracking-by-detection systems, various tracking algorithms can be employed, including Kalman filters, particle filters, and Intersection over Union (IoU) matching. All five object trackers examined in this paper (SORT ~\cite{bewley2016simple}, DeepOCSORT ~\cite{maggiolino2023deep}, OCSORT ~\cite{cao2023observation}, ByteTrack ~\cite{zhang2022bytetrack}, BoTSORT ~\cite{aharon2022bot}, and StrongSORT ~\cite{du2023strongsort}) utilize Kalman filters as their core tracking mechanism, with modifications to initial state parameters and other configuration settings to improve performance. These methods assume linear motion models that become increasingly inaccurate as object motion becomes more erratic, leading to the performance degradation we observe in our experiments.

Our research contributes to the field by providing a comprehensive analysis of Kalman filter-based tracking performance on fast-moving tiny objects. We address critical gaps in current tracking evaluation by introducing a novel racquetball dataset and specialized evaluation metrics. Specifically, we:
\begin{itemize}
    \item Present experimental results from implementing five state-of-the-art trackers on a novel racquetball dataset containing fast-moving tiny objects - the racquetball
    \item Analyze the performance characteristics of these trackers using ADE and AMD metrics and identify key areas for improvement
    \item Conduct thorough analysis of tracking limitations, identifying a critical trade-off where ByteTrack processes frames 15\% faster (26.6ms vs 26.8ms average) but exhibits 260\% higher tracking error compared to DeepOCSORT
    \item Demonstrate that current kalman filter-based approaches show limitations for tiny object tracking, with all methods exhibiting substantial spatial errors that make them unsuitable for precise tracking applications
    \item Provide detailed discussion of the architecture and working principles of Kalman filter and all five implemented tracking methods.
\end{itemize}

\section{Related Works}

\subsection{Overview of Object Tracking}
Visual object tracking has evolved significantly over the past decades, transitioning from traditional computer vision methods to sophisticated deep learning approaches visual object tracking is an important area in computer vision, and many tracking algorithms have been proposed with promising results. Modern tracking approaches can be broadly categorized into generative trackers~\cite{GenTracker}, discriminative trackers~\cite{Discriminative1}, ~\cite{Discriminative2}, ~\cite{Discriminative3}, ~\cite{Discriminative4}, and collaborative trackers ~\cite{collaborative1, collaborative2}, with deep neural network-based methods gaining prominence due to their outstanding performance. Recently, object tracking algorithms based on deep neural networks have emerged and obtained great attention from researchers due to their outstanding tracking performance.

The field has primarily focused on two main paradigms: tracking-by-detection and tracking-without-detection. In tracking-by-detection approaches, object detection is performed in each frame, followed by data association to link detections across frames. This methodology has proven effective for general object tracking scenarios but faces unique challenges when applied to tiny objects with rapid motion.

\subsection{Small Object Detection and Tracking}
Small object detection and tracking represent particularly challenging areas within computer vision Object detection and tracking are vital in computer vision and visual surveillance, allowing for the detection, recognition, and subsequent tracking of objects within images or video sequences. The ~\cite{surveySOD} survey paper highlights various approaches for small object detection and various benchmarks. The challenges arise from multiple factors including low resolution, limited discriminative features, and susceptibility to noise. Small object detection (SOD) is significant for many real-world applications, including criminal investigation, autonomous driving and remote sensing images. SOD has been one of the most challenging tasks in computer vision due to its low resolution and noise representation.

Recent research ~\cite{visDrone}, ~\cite{DOTA} has specifically addressed tiny object tracking challenges, with various studies creating specialized datasets for evaluation. However, most existing datasets focus on relatively static tiny objects or objects with predictable motion patterns. Despite these advances, tracking tiny objects with highly erratic and non-linear motion patterns, such as those found in fast-paced sports environments, remains an open challenge with limited available datasets.

\subsection{Sports Ball Tracking Applications}
Sports ball tracking has emerged as a specialized application domain with unique challenges and commercial significance. Ball tracking is crucial for AI systems to analyze sports effectively, but it's challenging due to factors like the ball's small size, high velocity, complex backgrounds, similar-looking objects, and varying lighting. Commercial systems like Hawk-Eye have demonstrated the practical value of accurate ball tracking. Hawk-Eye is a computer vision system used to visually track the trajectory of a ball and display a profile of its statistically most likely path as a moving image. It is used in more than 20 major sports.

Recent academic research has explored various approaches to sports ball tracking. For instance, research on table tennis ball tracking ~\cite{tracknet}, ~\cite{sun2020tracknetv2}, ~\cite{ju2020trajectory} and golf ball tracking ~\cite{golfball} has addressed similar challenges to those encountered in racquetball. In table tennis, due to the small size and rapid motion of the ball, identifying and tracking the table tennis ball through video is a particularly arduous task, where the majority of existing detection and tracking algorithms struggle to meet the practical application requirements. Tennis ball tracking systems have also been developed for robotic applications. The robotic vision system, installed on the tennis playing robot, utilized a trained neural network to detect tennis ball and predict ball trajectory.

However, most existing sports ball tracking research focuses on sports with more predictable ball trajectories, such as tennis or table tennis. Racquetball presents unique challenges due to its enclosed court environment, multiple wall interactions, and highly erratic ball motion patterns.

\subsection{Kalman Filter-Based Tracking Methods}
Kalman filter-based tracking has remained a cornerstone of object tracking systems due to its computational efficiency and theoretical foundation. The SORT ~\cite{bewley2016simple} (Simple Online and Realtime Tracking) algorithm established the baseline for modern KF-based tracking by combining Kalman filters with Hungarian algorithm ~\cite{kuhn1955hungarian} for data association. Subsequent developments including (DeepOCSORT~\cite{maggiolino2023deep}, OCSORT~\cite{cao2023observation}, ByteTrack~\cite{zhang2022bytetrack}, BoTSORT~\cite{aharon2022bot}, and StrongSORT~\cite{du2023strongsort}) have extended this foundation with various improvements.

OCSORT~\cite{cao2023observation} enhanced the original SORT algorithm by incorporating observation-centric innovations, while DeepOCSORT~\cite{maggiolino2023deep} further improved performance by integrating deep appearance features. ByteTrack~\cite{zhang2022bytetrack} focused on optimizing computational efficiency through dual-stage association strategies, and BoTSORT~\cite{aharon2022bot} combined motion and appearance modeling. StrongSORT~\cite{du2023strongsort} emphasized robust feature extraction and matching.

Despite these advances, Kalman filter-based methods rely on linear motion assumptions that may not adequately capture the complex dynamics of rapidly moving objects with frequent direction changes, such as balls in racquetball scenarios.

\subsection{Evaluation Metrics and Performance Assessment}

Traditional tracking evaluation has relied on metrics such as Multiple Object Tracking Accuracy (MOTA) and Multiple Object Tracking Precision (MOTP) for multi-object scenarios, as formalized in~\cite{bernardin2008evaluating} and widely adopted in benchmarks such as MOTChallenge~\cite{ristani2016performance} . However, these metrics may not fully capture the specific challenges associated with tiny object tracking, where precise localization accuracy becomes critical. Recent studies~\cite{luiten2021hota, li2009learning} have highlighted that MOTA and MOTP, while effective for multi-object settings, are less suitable for evaluating the trajectory tracking of a single object or for scenarios where fine-grained localization is paramount . Consequently, for our focus on single-object trajectory tracking—particularly involving tiny objects—alternative or more specialized evaluation metrics are necessary.

Our work employs Average Displacement Error (ADE) and Average Mahalanobis Distance (AMD) as evaluation metrics, focusing on trajectory accuracy rather than traditional identity-based metrics. This approach aligns with the specific requirements of tiny object tracking applications where maintaining precise spatial accuracy is paramount. Definition and more details about both ADE and AMD metrics can be found in ~\cite{ADE_metrics} and ~\cite{mohamed2022social} respectively.

\subsection{Research Gaps and Motivation}
While significant progress has been made in general object tracking and sports ball tracking, several research gaps remain:
\begin{itemize}
    \item Limited datasets: Most existing datasets focus on larger objects or sports with predictable ball motion patterns
    \item Evaluation metrics: Traditional metrics may not adequately assess tiny object tracking performance
    \item Motion modeling: Current KF-based methods assume linear motion, which is inadequate for highly erratic movement patterns
    \item Real-time performance: Balancing accuracy and speed remains challenging for tiny object tracking applications
\end{itemize}

Our work addresses these gaps by providing a comprehensive evaluation of state-of-the-art KF-based trackers on racquetball sequences, a sport characterized by highly erratic ball motion patterns. This evaluation reveals fundamental limitations of current approaches and highlights the need for specialized tracking methodologies for fast-moving tiny objects.

\section{Experimental Results and Analysis}
We conducted a comprehensive evaluation of object tracking algorithm performance on our custom racquetball dataset. The following subsections describe our dataset creation process, evaluation metrics, experimental results, and detailed performance analysis.

\subsection{Description of Dataset}
To evaluate the performance of five object trackers on tiny racquetball tracking tasks, we created a custom dataset to address the lack of publicly available datasets suitable for our research objectives. We used a ZED 2i stereo camera system to collect video sequences, though we utilized only the RGB frames for training our object detection model, despite the availability of depth information from the stereo setup.

\begin{figure}[htp]
    \centering
    \includegraphics[width=\linewidth]{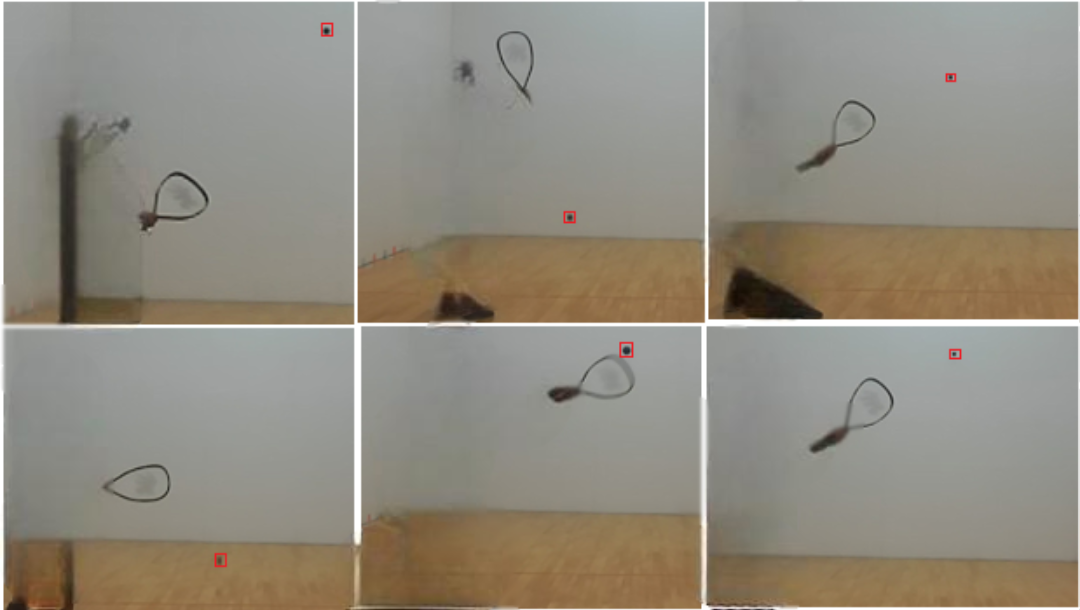}
    \caption{\centering Sample of our labeled dataset used for training. The person in the frames are removed for privacy protection. We also zoomed in to properly show the bounding box annotation of the ball.}
    \label{labeldataset}
\end{figure}

Our dataset consists of 10,000 frames captured at both 720p and 1280p resolutions, all carefully annotated to identify tiny ball locations. We divided the labeled dataset using a standard 70-20-10 split for training, validation, and testing respectively. Figure ~\ref{labeldataset} shows representative samples from our labeled dataset. Note that individuals have been removed from the displayed frames to protect privacy, although the complete dataset used for training and testing includes frames with people present.

The dataset release is currently pending review and approval from our Institutional Review Board (IRB). Upon receiving IRB approval, we plan to make the dataset publicly available through a dedicated GitHub repository to support future research in this area.

\subsection{Object Detection System}
We evaluated multiple well-established object detection methods including YOLOv5, YOLOv7, YOLOv8, and Faster-RCNN to identify the most effective approach for detecting tiny balls in our dataset. Through systematic comparison, we found that YOLO family detectors, particularly YOLOv5 and YOLOv8, demonstrated superior performance in detecting small ball objects with higher precision scores.
After analyzing both accuracy and inference time characteristics of YOLOv5 and YOLOv8, we selected trained YOLOv8 weights as the object detector for all five tracking methods (OCSORT, DeepOCSORT, StrongSORT, BoTSORT, and ByteTrack). Table ~\ref{ObjectDetection} presents the performance comparison of all YOLO models along with their respective inference speeds at the specified resolution.

\begin{table}[htp]
\centering
\begin{tblr}{
  row{2} = {c},
  cell{1}{6} = {c=2}{c},
  vline{2-7} = {2-5}{},
  vline{6-7} = {-}{},
  hline{2-3} = {-}{},
}
        &                &                &               &               & Inference Speed &              \\
Methods & Precision      & Recall         & mAP@0.5       & F1            & Latency (ms)    & FPS (approx) \\
YOLOv5  & \textbf{0.856} & \textbf{0.845} & \textbf{0.79} & \textbf{0.85} & 58.7            & 17           \\
YOLOv7  & 0.70           & 0.63           & 0.64          & 0.66          & 59              & 17           \\
YOLOv8  & 0.84           & 0.80           & 0.76          & 0.82          & \textbf{24.5}   & \textbf{41}  
\end{tblr} 
\newline
\newline
\caption{\centering This table shows the performance accuracy of YOLO family object detectors and their respective inferencing speed for image of 1280x720 resolution.}
\label{ObjectDetection}
\end{table}

\subsection{ Tracking Evaluation Metrics}
For comprehensive tracking performance evaluation of all five Kalman filter-based trackers, we employed Average Displacement Error (ADE) and Average Mahalanobis Distance (AMD) as our primary performance metrics. Both metrics are measured in pixels, where one pixel corresponds to approximately 0.026 cm in our experimental setup. \par

\textbf{Average Displacement Error (ADE)}: ADE calculates the mean Euclidean (L2) distance between predicted trajectory points and ground truth positions. It quantifies the overall spatial accuracy of tracking by summing displacement errors at each position and computing their average:
\[{ADE} = \frac{\sum ^{n}_{1} \sqrt{(x_{2}-x_{1})^2 + (y_{2}-y_{1})^2}}{N_{traj}}\]

\textbf{Average Mahalanobis Distance (AMD)}: AMD represents the mean Mahalanobis Distance between all corresponding point pairs in predicted and ground truth trajectories. Mahalanobis Distance measures the distance between pair points (predicted and ground truth) in multivariate space while considering correlations between variables and variable variances. For two vector points, $p_{pred}$ and $p_{truth}$, in p-dimensional space, Mahalanobis Distance is calculated as:
\[MD_{i}(\mathbf{p_{pred}}, \mathbf{p_{truth}}) = \sqrt{(\mathbf{p_{pred}} - \mathbf{p_{truth}})^\top \mathbf{\Sigma}^{-1} (\mathbf{p_{pred}} - \mathbf{p_{truth}})}\] 
Now for AMD, we sum all MD values and divide by the total number of point pairs N. 
\[ AMD = \frac{1}{N}\sum_{i=1}^{N}MD_{i}\]

\subsection{Tracking Performance Results and Analysis}
\begin{figure}[htp]
    \centering
    \includesvg[width=\textwidth]{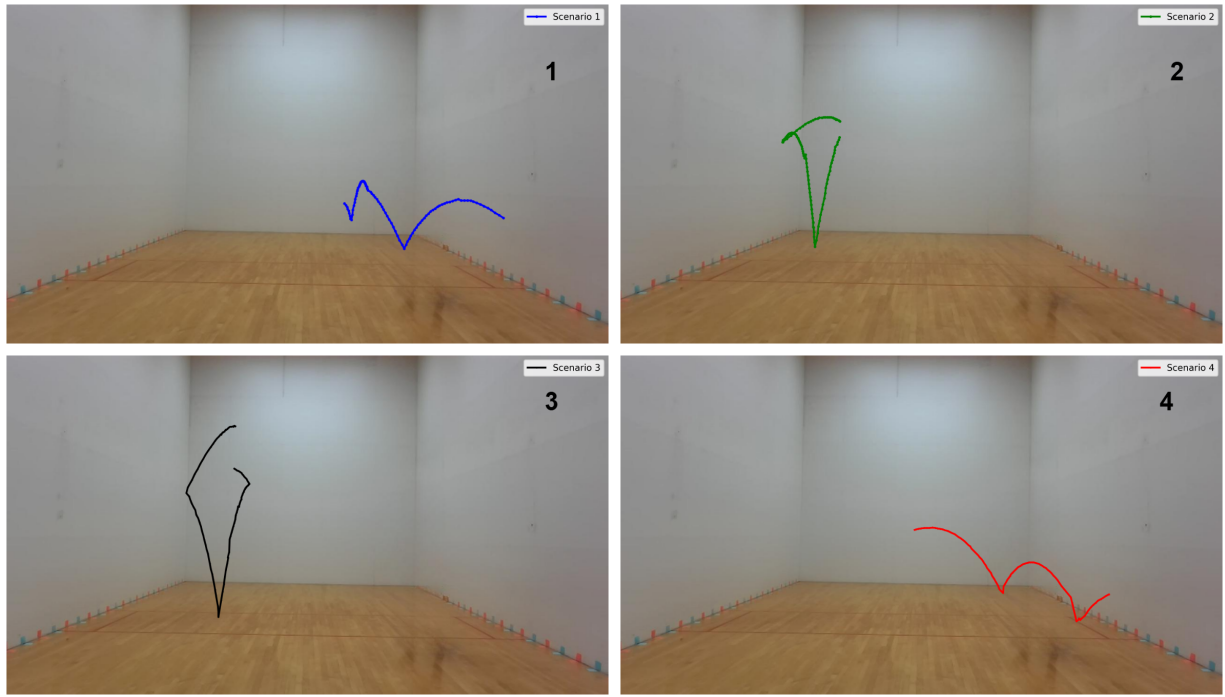}
    \caption{\centering Illustration of four scenarios we evaluated all the trackers on. The trajectory of the ball for each scenario is plotted on the top of actual training image to show real trajectory of the scenario.}
    \label{scenarios}
\end{figure}

We conducted tracking analysis across four distinct scenarios, with each scenario representing a brief but challenging instance of racquetball gameplay. Table ~\ref{OverallTracking} presents the performance results for all five trackers across these four scenarios, while Figure ~\ref{scenarios} illustrates the specific scenarios used in our evaluation.

Our objective in evaluating tracker performance across four distinct scenarios was to understand how each tracking method performs under different challenging conditions. This comprehensive analysis provides robust insights into the limitations and strengths of all five tracking approaches.

\definecolor{NavyBlue}{rgb}{0,0,0.545}
\begin{table}[htp]
\centering
\resizebox{0.85\linewidth}{!}{%
\begin{tblr}{
  cells = {c},
  cell{1}{2} = {c=2}{fg=NavyBlue},
  cell{1}{4} = {c=2}{fg=NavyBlue},
  cell{1}{6} = {c=2}{fg=NavyBlue},
  cell{1}{8} = {c=2}{fg=NavyBlue},
  vline{2-3,5,7} = {-}{},
  vline{3-4,5-6,7-9} = {-}{},
  vline{2-9} = {2-7}{},
  hline{2-3} = {-}{},
  hline{4-7} = {2-9}{},
}
                                         & \textbf{Scenario 1}     &               & \textbf{Scenario 2}     &               & \textbf{Scenario 3}     &               & \textbf{Scenario 4}    &               \\
Tracking Methods\textbackslash{} Metrics & ADE            & AMD           & ADE            & AMD           & ADE            & AMD           & ADE           & AMD           \\
DeepOCSORT                               & \textbf{22.97} & \textbf{0.69} & \textbf{31.31} & \textbf{0.91} & \textbf{50.42} & \textbf{0.91} & \textbf{19.9} & \textbf{0.87} \\
OCSORT                                   & 65.26          & 1.83          & 93.5           & 2.54          & 111.8          & 2.54          & 39.16         & 2.09          \\
StrongSORT                               & 99.82          & 2.86          & 169.7          & 23.9          & 157.22         & 23.9          & 153.7         & 2.84          \\
BoTSORT                                  & 51.7           & 1.81          & 158            & 11.59         & 74.11          & 11.59         & 178.5         & 1.67          \\
ByteTrack                                & 39.4           & 3.22          & 146.4          & 13.02         & 56.48          & 13.02         & 215           & 2.39          
\end{tblr} }
\newline
\newline
\caption{\centering This table shows ADE and AMD measurement for each predicted trajectory for each scenario. As seen, DeepOCSORT performs better in correctly tracking the trajectory of the tiny ball among all trackers.}
\label{OverallTracking}
\end{table}

The results clearly demonstrate that DeepOCSORT achieves the most accurate tracking performance among all tested methods, consistently showing the lowest ADE and AMD values across scenarios. However, all trackers—including DeepOCSORT, OCSORT, ByteTrack, BoTSORT, and StrongSORT—struggle significantly with the erratic motion patterns and rapid scale changes characteristic of fast-moving racquetballs, resulting in poor overall tracking performance and fragmented trajectory outputs.

We believe these performance limitations stem from the trackers' internal configuration and state assumption mechanisms, which may not be adequately designed to handle the complex scenarios encountered in tiny object tracking. Additionally, since these trackers operate on the tracking-by-detection principle, tracker failures often occur when the underlying object detector fails to detect the ball, creating a cascading effect that compromises overall tracking performance.

It is also important to note that all these trackers are designed based on Kalman filter (KF). The KF assumptions fail in tracking tiny objects like racquetball because tiny balls don't exhibit linear behavior for longer time. There is a short and frequent change in motion that breaks the internal state parameters of the KF leading to poor tracking performance. Tracking of other tiny objects like drones, space debris can still be done with precision using KF or KF-based trackers because drone and debris don't have erratic motion change or frequent collision with another object.

\subsubsection{Analysis on Inferencing and Update Frequency}
Two critical parameters that significantly impact tracking algorithm performance are inference speed and update frequency per image. Table ~\ref{all_scenarios_speed} and figure ~\ref{performance_analysis} presents detailed measurements of inference and update speeds for all five trackers across our four evaluation scenarios. 

\textbf{Inferencing speed}, also known as inference time or frames per second (FPS), this parameter measures how quickly the tracking algorithm processes each video frame. Inference speed encompasses the total time required to: (a) run the object detector on a frame to identify target objects, and (b) execute the tracker to associate detections with existing tracks and update internal object states. Higher inference speeds enable processing of more frames per second, which is crucial for real-time applications. Multiple factors influence inference speed, including object detector complexity (single-stage versus multi-stage detectors), computational efficiency of the tracking algorithm, and the hardware platform used for implementation. 

\textbf{Update frequency per image} this parameter measures how frequently the tracker updates its internal state using new observations from each video frame to account for changes in object appearance or location. For every processed frame, the tracker detects objects, associates them with existing tracks using prediction and matching algorithms, and updates the position, appearance, and state information for each tracked object based on new detections. The update process helps maintain accurate object identities in multi-object tracking scenarios, handles occlusions effectively, and recovers from tracking drift or missed detections.

\begin{table} [!ht]
\centering
\resizebox{0.85\linewidth}{!}{%
\begin{tblr}{
  cells = {c},
  row{1} = {fg=NavyBlue},
  cell{2}{1} = {r=2}{},
  cell{4}{1} = {r=2}{},
  cell{6}{1} = {r=2}{},
  cell{8}{1} = {r=2}{},
  vline{3-7} = {-}{},
  hline{2,4,6,8} = {-}{},
  hline{3,5,7,9} = {3-7}{dashed},
}
\textbf{Scenario} & \textbf{Metric}   & \textbf{DeepOCSORT} & \textbf{OCSORT} & \textbf{StrongSORT} & \textbf{BOTSORT} & \textbf{Bytetrack} \\
1                 & Inferencing speed & 31.1~               & 30.8~           & 31.1                & 32.0~            & 32.5               \\
                  & Update per image  & 61.5~               & 9.6~            & 39.5~               & 60.5             & 1.4                \\
2                 & Inferencing speed & 24.8~               & 24.7~           & 24.9~               & 24.8             & 24.7               \\
                  & Update per image  & 34.1~               & 4.9~            & 24.4                & 34.7             & 0.6                \\
3                 & Inferencing speed & 24.9~               & 24.7~           & 25.4~               & 25.3             & 24.7               \\
                  & Update per image  & 36.9~               & 4.4             & 24.5~               & 36.0             & 0.6                \\
4                 & Inferencing speed & 26.2                & 24.8            & 25.3                & 26.5             & 24.6               \\
                  & Update per image  & 34.5                & 6.5             & 27.1                & 39.0             & 0.6                
\end{tblr} }
\newline
\newline
\caption{\centering Inferencing and update speed per image across four scenarios for all trackers. From the table we can see that ByteTrack on average has better inferencing and internal update speed on four evaluated scenarios. Both inferencing and update per image are measured in milliseconds (ms).}
\label{all_scenarios_speed}
\end{table}

A significant trade-off exists between frequent updates and tracking accuracy. While frequent updates enhance adaptability to rapid changes, they also increase the risk of model drift when dealing with missed or false detections. Conversely, infrequent updates improve tracking stability and accuracy but reduce responsiveness to sustained appearance changes. Confidence-based update strategies often provide optimal performance by balancing these competing requirements.

Inference speed and update frequency demonstrate an inverse relationship: higher update frequencies typically reduce inference speeds due to increased computational overhead from tasks such as feature extraction, motion prediction, and internal state updates.

ByteTrack exemplifies this trade-off by achieving good real-time performance through optimized update logic. It employs a dual-stage association strategy with high and low confidence detections combined with lightweight Kalman filters, minimizing computational overhead while maintaining reasonable tracking accuracy. This design results in ByteTrack achieving an average inference speed of 26.6ms and update frequency of 0.8ms averaged across all scenarios. However, as shown in Table ~\ref{OverallTracking}, ByteTrack exhibits considerably higher ADE and AMD values (average of 114 and 6.8 respectively), indicating reduced tracking accuracy.

In contrast, StrongSORT implements complex update policies with deep feature embeddings that reduce processing speed but potentially improve tracking precision under certain conditions. OCSORT represents a balanced approach, managing the trade-off between inference and update speeds through confidence-based updates that trigger internal state refresh only when detection reliability exceeds predetermined thresholds. This strategy preserves processing speed without significantly sacrificing tracking robustness. OCSORT achieves average ADE and AMD scores of 79.9 and 1.84 respectively, with average inference and update speeds of 26.25ms and 6.35ms.

\begin{figure}[htbp]
    \centering
    \includesvg[width=\linewidth]{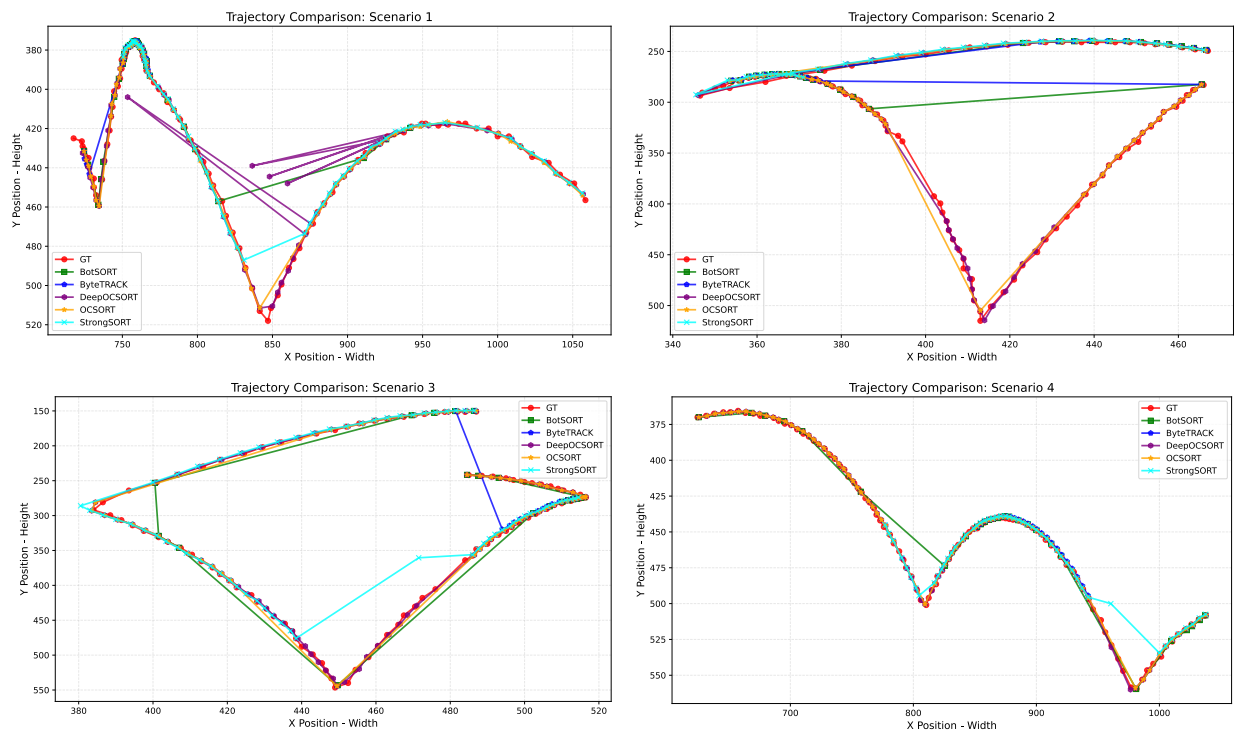}
    \caption{\centering 2D plot comparing trajectories of all four scenario based on results of five trackers evaluated. The points outside of normal trajectory path represent tracking drift leading to fragmented trajectory and higher ADE and AMD error value.}
    \label{scenarioTraj}
\end{figure}

However, the erratic and non-linear motion patterns typical of racquetball — including multiple bounces and collisions with sidewalls, motion induced blur — present unique challenges that faster update frequencies alone cannot address effectively. Figure ~\ref{scenarioTraj} illustrates the trajectory comparisons for each scenario across all tracking methods relative to ground truth trajectories. The visualization clearly shows that several trackers lose track of the ball for extended periods, while others exhibit significant tracking drift due to poor detection quality, resulting in fragmented trajectory outputs.

\begin{figure}[htp]
    \centering
    \includesvg[width=\linewidth]{tracking_performance_analysisV2}
    \caption{Performance comparison of tracking methods. Top row: Speed analysis showing (a) inference time and (b) update time per image. Middle row: Accuracy analysis displaying (c) Average Displacement Error (ADE) and (d) Average Mahalanobis Distance (AMD). Bottom row: Trade-off analysis between (e) speed metrics and (f) accuracy metrics. Lower values indicate better performance in all metrics. Bytetrack demonstrates fastest update speeds while DeepOCSORT achieves best accuracy scores. Data averaged across four video scenarios.}
    \label{performance_analysis}
\end{figure}

\section{Future Research Direction}
Work in this paper focuses exclusively on evaluating five state-of-the-art Kalman filter-based tracking methods using our novel racquetball dataset. We have demonstrated the performance characteristics of these trackers when applied to tiny object tracking scenarios. As our results indicate, these five tracking algorithms show limited reliability when tracking small, fast-moving objects with unpredictable motion patterns.
Future research should prioritize the development of tracking algorithms capable of handling complex ball behavior in real-time applications. Potential research directions include creating hybrid tracking systems that combine enhanced object detection methods with tracking algorithms specifically designed to account for the complex movement patterns and spatial characteristics of small, fast-moving objects. Additional work should examine the specific limitations of current Kalman filter-based approaches and explore how these systems can be improved (like hyperparameter tuning) or replaced with more suitable alternatives for tiny object tracking scenarios. A physics inspired tracking framework where we can model the complex behavior of the ball might be great avenue to explore. A newer object detection model with infused physics as a trainable parameter is other avenue to explore.

\section{Conclusion}
We paper presents comprehensive experimental results evaluating five state-of-the-art tracking algorithms across four unique scenarios from our custom racquetball dataset. We implemented and analyzed the performance of DeepOCSORT, OCSORT, ByteTrack, BoTSORT, and StrongSORT on a novel dataset specifically designed for fast-moving tiny object tracking applications.

Our results demonstrate that these five Kalman filter-based trackers exhibit poor performance when tracking fast-moving small objects due to the erratic and non-linear motion patterns characteristic of such objects. We conducted detailed analysis of two key performance parameters—inference speed and update frequency per image—that significantly impact tracker effectiveness. Our analysis reveals how these parameters influence both tracking accuracy and overall processing speed.

Our findings indicate that current Kalman filter-based tracking approaches are inadequate for reliable tiny object tracking applications, particularly in scenarios involving unpredictable motion patterns like those found in racquetball gameplay. This research highlights the need for developing new tracking methodologies or significantly improving existing approaches to achieve robust and reliable tracking of fast-moving tiny objects in real-world applications.
            

\section{Review of Trackers and Kalman Filter}
In this section, we provide comprehensive review about the core features of all five trackers as well as mathematical formulation of Kalman filter. We also give overview about SORT algorithm which lays foundation for all other trackers discussed. 

\subsection{Simple Online and Realtime Tracking (SORT)}

The Simple Online and Realtime Tracking (SORT) ~\cite{bewley2016simple} algorithm offers a pragmatic approach to multiple object tracking, emphasizing efficient object association for online and realtime applications. Detection quality is crucial, with improvements in the detector leading to significant gains in tracking performance. SORT relies on a lean implementation of the tracking-by-detection framework, processing detections in each frame as bounding boxes.

SORT uses the Kalman Filter to predict object motion and the Hungarian algorithm~\cite{kuhn1955hungarian} for data association, achieving accuracy comparable to state-of-the-art online trackers while running at 260 Hz, significantly faster than many alternatives. Appearance features are ignored beyond the detection component, focusing solely on bounding box position and size for motion estimation and data association. Short-term and long-term occlusions are also ignored to reduce complexity. The algorithm's design facilitates efficient handling of common frame-to-frame associations.

\begin{figure}
    \centering
    \includegraphics[width=0.9\linewidth]{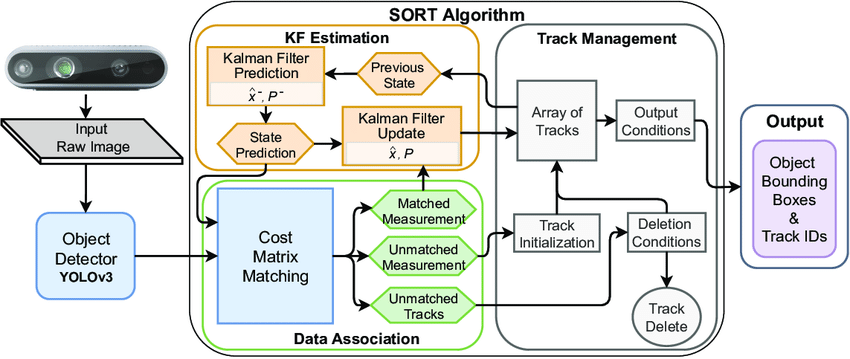}
    \caption{Overview of SORT Algorithm for object tracking ~\cite{SORT_Pic}.}
    \label{SORT_Pic}
\end{figure}

The object state is modeled using horizontal and vertical pixel locations, scale, and aspect ratio, with inter-frame displacements approximated by a linear constant velocity model. When a detection is associated with a target, the detected bounding box updates the target state via a Kalman filter. The assignment cost matrix is computed using intersection-over-union (IOU) distance between detections and predicted bounding boxes, solved optimally with the Hungarian algorithm. New identities are created for objects entering the image, undergoing a probationary period to prevent false positives, and terminated if not detected for a set number of frames. Figure ~\ref{SORT_Pic} shows general outline as presented in ~\cite{SORT_Pic}.

\subsection{ByteTrack}

ByteTrack~\cite{zhang2022bytetrack}, introduced by Zhang et al. in 2022, is a multi-object tracking (MOT) method designed to mitigate issues of object loss and fragmented trajectories common in trackers that discard detections with low confidence scores. Unlike traditional methods, ByteTrack associates every detection with tracklets, not just high-confidence ones. For low-score detection boxes, ByteTrack leverages similarities with existing tracklets to differentiate true objects from background noise.

In its original implementation, ByteTrack employs YOLOX as an object detector, with the BYTE algorithm associating detected bounding boxes across frames. A Kalman filter predicts object locations in subsequent frames, and the tracker uses IoU (Intersection over Union) or Re-ID feature similarity to identify and maintain tracks. This approach proves particularly effective in scenarios where occlusion and blurriness might compromise tracking performance.

ByteTrack utilizes a tracklet interpolation technique to estimate object positions when tracklets are temporarily lost due to occlusion or size changes. By considering almost every detection box and using tracklet interpolation, ByteTrack has better tracking capabilities for difficult to detect and track objects. The method has demonstrated notable improvements across various MOT benchmarks and can be broadly applied to other tackers. 

Suppose we have missing tracklets from frame $t_{1}$ to $t_{2}$ due to some reason like blurriness, or occlusion. In that case, the tracklet interpolation technique can help improve tracking and performance scores. Suppose $B_{t}$ represents the bounding box coordinates [x,y,w,h] of the tracklets, where x is the top-left coordinate and y is the bottom-right coordinate of the bounding box. Given that we have a threshold set, we can use the interpolation technique, as in equation 1, to interpolate the box coordinates of frame t. Refer to the original paper ~\cite{zhang2022bytetrack} and implementation code on GitHub ~\cite{github}. 
\begin{equation}
    B_t = B_{t_{1}} + (B_{t_{2}}-B_{t_{1}})\frac{t-t_1}{t_2-t_1}, t_1< t< t_2
\end{equation}

\begin{figure}[htp]
    \centering
    \includesvg[width=0.9\textwidth]{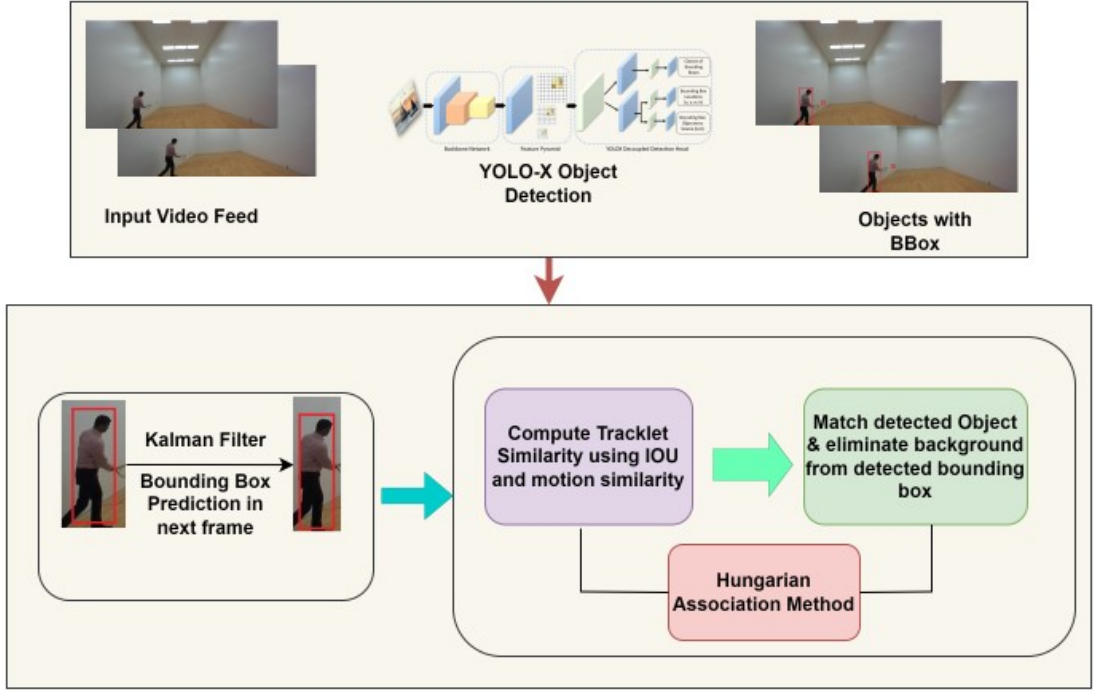}
    \caption{Outline of working of ByteTrack Tracker.}
    \label{ByteTrack}
\end{figure}

\subsection{BoT-SORT: Robust Associations Multi-Pedestrian Tracking}

BoT-SORT ~\cite{aharon2022bot} is a robust, state-of-the-art tracker designed for multi-object tracking (MOT), which aims to detect and track objects in a scene while maintaining unique identifiers for each object. It enhances tracking-by-detection methods by integrating motion and appearance information, camera-motion compensation (CMC), and a refined Kalman Filter (KF) state vector. The tracker's architecture builds upon ByteTrack ~\cite{zhang2022bytetrack} and addresses limitations found in "SORT-like" algorithms.

One key feature of BoT-SORT is its camera motion compensation, which addresses the problem of camera motion that can reduce the overlap between predicted tracklet bounding boxes and detected bounding boxes. By using image registration techniques, BoT-SORT estimates camera motion and corrects the Kalman filter, improving tracking performance in dynamic camera situations.

The Kalman Filter is another important aspect of BoT-SORT, which is used to model the object's motion in the image plane. BoT-SORT modifies the KF's state vector to directly estimate the width and height of bounding boxes, leading to more accurate localization. The method also uses a new approach for fusing IoU (Intersection over Union) and ReID (Re-Identification) cosine distance for robust associations between detections and tracklets. This fusion method rejects low cosine similarity candidates and uses the minimum value from the IoU distance matrix and the cosine distance matrix to construct the cost matrix. BoT-SORT's design and features lead to its high ranking in MOTA, IDF1, and HOTA metrics on MOT17 and MOT20 datasets. 

\begin{figure}[htp]
    \centering
    \includesvg[width=\textwidth]{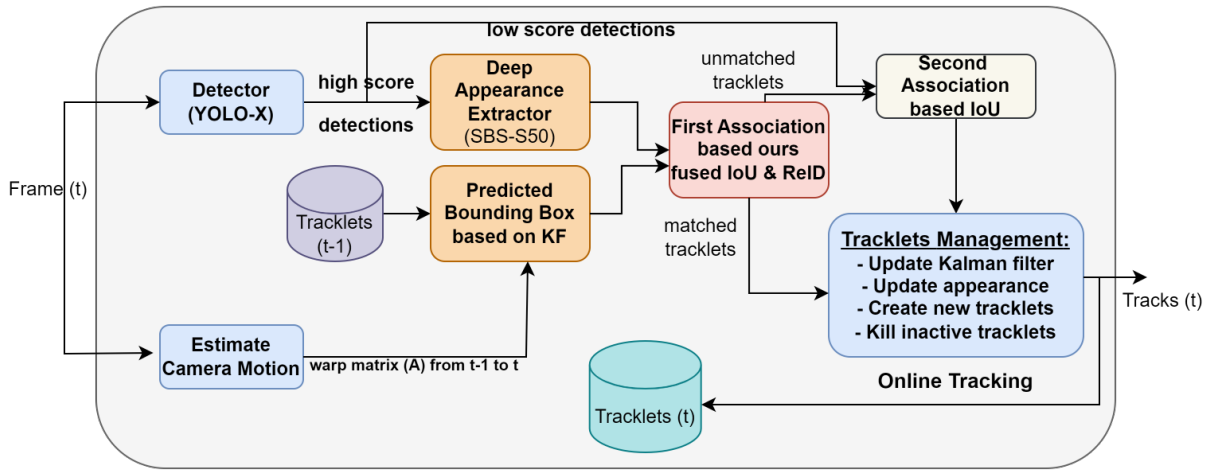}
    \caption{Architectural Overview of BoT-SORT algorithm ~\cite{aharon2022bot}.}
    \label{BoTSORT}
\end{figure}

The initial state vector $x_{k}$ is represented as equation 1, and the measurement vector $z_{k}$ is shown as equation 2. The noise covariance matrix, $Q_k$, and measurement covariance $R_k$ will also be modified to include the width and height of the bounding box for state estimation. Like in ByteTrack, linear interpolation is used to interpolate positions of missing tracklets over time. The figure ~\ref{BoTSORT} represents the original architecture of the BoT-SORT. 
\begin{equation}
    x_k = \begin{bmatrix}
x_c(k) &y_c(k)  &w(k)  &h(k)  &\dot{x}_c(k)  & \dot{y}_c(k) &\dot{w}_c(k)  & \dot{h}_c(k)
\end{bmatrix}^T
\end{equation}

\begin{equation}
    z_k = \begin{bmatrix}
    z_{x_{c}}(k) &z_{y_{c}}(k)  & z_{w_{c}}(k) & z_{h_{c}}(k)
    \end{bmatrix}^T
\end{equation}

\subsubsection{Camera Motion Compensation}
Camera Motion Compensation (CMC) in BoT-SORT addresses issues caused by camera movement, which can reduce the overlap between predicted tracklet bounding boxes and detected bounding boxes. This is particularly important in dynamic camera situations where bounding box locations can shift dramatically, leading to ID switches or false negatives.

CMC estimates the camera's rigid motion by using image registration between adjacent frames. It employs a global motion compensation (GMC) technique, similar to that used in OpenCV's Video Stabilization module, with affine transformation. The process begins with extracting image keypoints followed by sparse optical flow for feature tracking with translation-based local outlier rejection. RANSAC ~\cite{fischler1981random} is then used to solve for the affine matrix $A_{k-1}^{k}$.

The calculated affine matrix $A_{k-1}^{k}$ transforms the predicted bounding box from the coordinate system of frame k-1 to frame k. The translation part of the transformation matrix affects the center location of the bounding box, while the other part affects the entire state vector and noise matrix. The camera motion correction is applied through specific equations involving matrices M and T, which represent the scale/rotations and translation parts of the affine matrix, respectively. These equations update the Kalman Filter's predicted state vector and covariance matrix to compensate for camera motion before the Kalman Filter update step. By compensating for rigid camera motion, BoT-SORT maintains accurate tracking even when the camera is moving.

\subsection{Observation-Centric SORT (OC-SORT)}
\begin{figure}[htp]
    \centering
    \includegraphics[width=\linewidth]{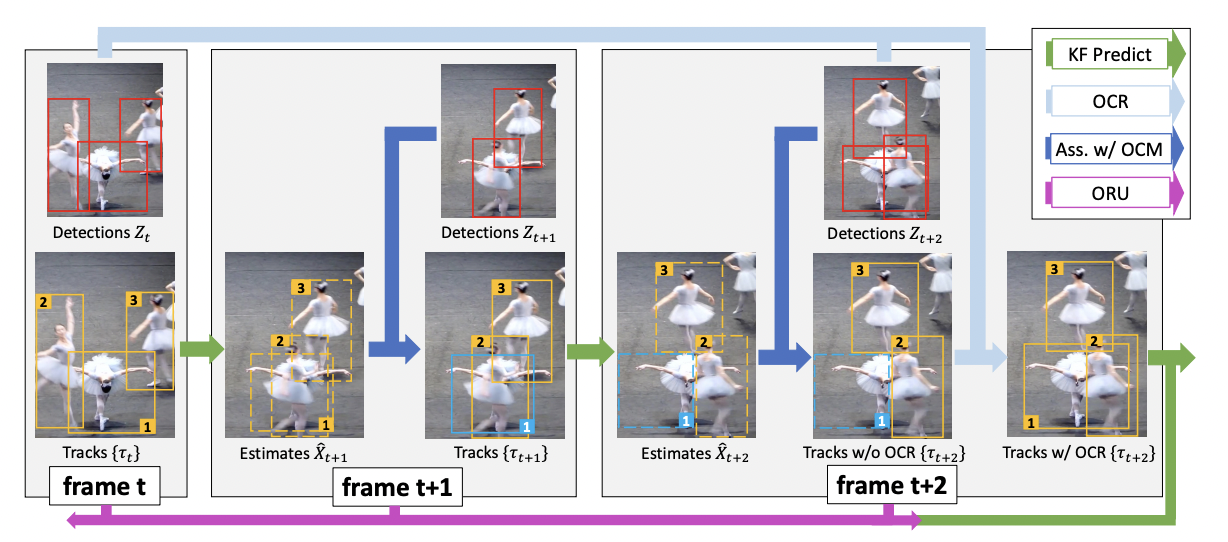}
    \caption{Overview of a general pipeline of OCSORT algorithm as shown in\cite{cao2023observation}.}
    \label{OC-SORT}
\end{figure}

OC-SORT ~\cite{cao2023observation} is a pure-motion-model-based multi-object tracker (MOT) which aims to increase tracking robustness in crowded scenarios (occulsion and non-linear motion). It addresses limitations in the original SORT algorithm by incorporating object observations to correct errors accumulated during occlusion, rather than relying solely on linear state estimations. OC-SORT maintains simplicity, online processing, and real-time performance, while significantly improving robustness. The figure ~\ref{OC-SORT} represents the pipeline of the OCSORT algorithm \cite{cao2023observation}. 

The method works by re-evaluating the standard Kalman filter (KF) and addressing its shortcomings, such as sensitivity to state estimation noise, error accumulation over time, and an estimation-centric approach. OC-SORT introduces two main innovations: Observation-Centric Re-Update (ORU) and Observation-Centric Momentum (OCM). The pipeline of OC-SORT also includes Observation-Centric Recovery (OCR) technique.

\textbf{Observation-Centric Re-Update (ORU)} reduces accumulated error during periods when a track is lost due to occlusion. When a track is re-activated, ORU backchecks the period it was untracked and re-updates the KF parameters based on a virtual trajectory generated from the last-seen observation and the re-activating observation. This virtual trajectory provides "observations" for a re-update loop, correcting the error accumulation from the dummy update process.

\textbf{Observation-Centric Momentum (OCM)} incorporates direction consistency of tracks in the cost matrix for association. OCM uses object state observations to calculate a motion direction, which helps in the association process by adding a velocity consistency cost. This approach reduces noise in motion direction calculation.

\textbf{Observation-Centric Recovery (OCR)} is a heuristic technique applied to recover tracks from being lost. OCR initiates a second attempt at associating unmatched tracks with unmatched observations after the initial association stage, useful when an object stops or is briefly occluded. In our case, the velocity of the ball changes as it moves through space and bounces off the surface so OC-SORT as a tracker is effective. 

\subsection{Deep OC-SORT}
Introduced by Gerard Maggiolino and et.al in ~\cite{maggiolino2023deep}, Deep OC-SORT is a direct extension of OC-SORT with the addition of appearance cues to the tracker module. It introduces a new way to adaptively integrate object appearance matching into motion-based methods. It leverages visual cues to improve tracking robustness, especially in scenarios with occlusion, motion blur, or similar object appearances. Deep OC-SORT achieves state-of-the-art performance on MOT17, MOT20, and DanceTrack benchmarks. 

Deep OC-SORT comprises three main modules: Camera Motion Compensation (CMC), Dynamic Appearance (DA), and Adaptive Weighting (AW). CMC refines object localization by compensating for camera motion between frames, applying scaled rotation and translation to OC-SORT's components. For Observation Centric Momentum (OCM) and CMC, the bounding box corner points are adjusted to account for camera motion. Similarly, for Observation Centric Recovery (OCR) and CMC, the bounding box position is adjusted to correct its position under CMC.

\textbf{The Dynamic Appearance (DA)} module adaptively incorporates visual information by adjusting the weighting factor $\alpha$ in the Exponential Moving Average (EMA) based on detector confidence. This allows the system to selectively use appearance information in high-quality detections and ignore corrupted embeddings from occlusions or blur. The standard EMA is modified with a changing $\alpha$t, which depends on the detector confidence score (sdet) and a confidence threshold ($\alpha$).

\textbf{Adaptive Weighting (AW)} boosts the weight of appearance features based on their discriminativeness. It increases the appearance weight over track-box scores based on how distinctly a track is associated with a single detection, or vice versa. The discriminativeness is measured using the difference between the highest and second-highest similarity scores (zdiff). This results in a final cost matrix that combines IoU and appearance information, enhancing the matching process. Deep OC-SORT improves upon OC-SORT through these enhancements, leading to more accurate and robust multi-object tracking. The outline of Deep OC-SORT is shown below in Figure ~\ref{DeepOCSORT}. 

\begin{figure}[htp]
    \centering
    \includegraphics{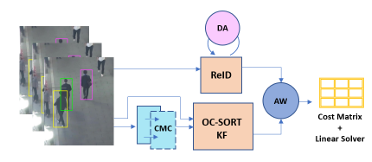}
    \caption{Illustration of Deep OC-SORT algorithm as shown in~\cite{maggiolino2023deep}. }
    \label{DeepOCSORT}
\end{figure}
 
\subsection{Strong-SORT}
\begin{figure}
    \centering
    \includesvg[width = 0.7\textwidth]{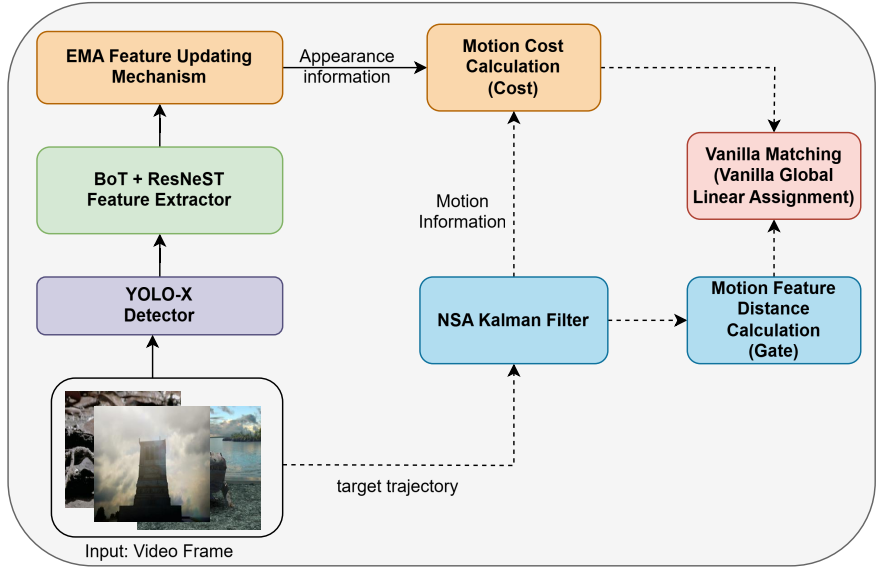}
    \caption{Framework of StrongSORT redesigned based on the original paper.}
    \label{StrongSort}
\end{figure}

StrongSORT~\cite{du2023strongsort} is a multi-object tracking (MOT) baseline developed to enhance the classic DeepSORT~\cite{wojke2017simple} tracker by integrating advanced modules and inference techniques. It addresses the challenges of fair comparisons in MOT research, where varying detectors, ReID models, and training tricks complicate performance assessment. StrongSORT improves upon DeepSORT by incorporating a stronger detector (YOLOX-X) and a more powerful appearance feature extractor (BoT).Figure ~\ref{StrongSort} shows general outline of the StrongSORT.

\subsubsection{Key Features and Methods}
\begin{itemize}
    \item \textbf{Advanced Modules}: StrongSORT replaces DeepSORT's original detector with YOLOX-X for improved detection performance. It also utilizes BoT, a stronger appearance feature extractor, to obtain more discriminative features.
    \item \textbf{EMA Feature Updating}: StrongSORT employs an exponential moving average (EMA) to update appearance features, reducing sensitivity to detection noise and enhancing matching quality. The EMA updating leverages inter-frame feature changes to improve tracking robustness.
    \item \textbf{ECC for Camera Motion Compensation}: To handle camera movements, StrongSORT incorporates the enhanced correlation coefficient maximization (ECC) model. ECC estimates global rotation and translation between adjacent frames, compensating for motion noise caused by camera movement.
    \item \textbf{NSA Kalman Filter}: StrongSORT adopts the NSA Kalman filter to address the limitations of the vanilla Kalman filter, which is vulnerable to low-quality detections. The NSA Kalman filter adaptively calculates noise covariance based on detection confidence scores, improving the accuracy of updated states.
    \item \textbf{AFLink}: Appearance-Free Link (AFLink) addresses missing association by using only spatiotemporal information to predict the connectivity between tracklets, balancing speed and accuracy. AFLink uses a two-branch framework, and only requires 10 seconds for training and 10 seconds for testing on MOT17.
    \item \textbf{GSI}: Gaussian-smoothed Interpolation (GSI) compensates for missing detections by modeling nonlinear motion using Gaussian process regression, which refines interpolated localizations without additional time-consuming components. GSI serves as a detection noise filter, producing stable localizations beneficial for speed estimation and related tasks.
    \item \textbf{Vanilla Matching}: StrongSORT employs vanilla matching, a global linear assignment approach, to solve the association problem between detected objects and existing tracks. Unlike the matching cascade algorithm used in DeepSORT, which prioritizes frequently seen objects, vanilla matching considers all possible pairings of detections and tracks simultaneously, aiming for the optimal assignment based on a cost matrix that integrates both appearance and motion information. The authors found that as the tracker becomes stronger, the matching cascade's prior constraints can limit performance; thus, vanilla matching improves matching accuracy by better exploiting appearance and motion cues, leading to gains such as a 1.4 increase in IDF1 for StrongSORTv5.
\end{itemize}

\subsection{Kalman Filter: Mathematical Foundations and Algorithm}
\subsubsection{Introduction}
The Kalman Filter (KF)~\cite{kalman1960new} is a recursive state estimation algorithm that provides an optimal solution to the linear filtering problem in the presence of Gaussian noise. Named after Rudolf E. Kálmán, this algorithm has become fundamental in various fields including navigation, control systems, signal processing, and computer vision - particularly for object tracking applications.
The power of the Kalman Filter lies in its ability to:
\begin{itemize}
\item Provide estimates of the state of a dynamic system from noisy measurements
\item Incorporate prior knowledge about system dynamics
\item Account for uncertainty in both measurements and system dynamics
\item Update estimates recursively, making it computationally efficient
\item Provide not just state estimates but also uncertainty estimates
\end{itemize}

\subsubsection{Mathematical Formulation}
We base our mathematical formulation of Kalman Filter based on the '\textit{Probabilistic Robotics}' ~\cite{thrun2002probabilistic} by Sebastian Thrun. The Kalman Filter works by maintaining two probabilistic models:
\begin{enumerate}
\item A \textbf{process model} that describes how the state evolves over time
\item A \textbf{measurement model} that relates the measurements to the state
\end{enumerate}
Both models incorporate Gaussian noise, and the filter maintains the state as a multivariate Gaussian probability distribution characterized by a mean vector and a covariance matrix.

\subsubsubsection{System Model}
The Kalman Filter assumes a linear dynamic system represented by: \newline
\textbf{State transition equation:}
\begin{equation}
\mathbf{x}_k = \mathbf{F}_k\mathbf{x}_{k-1} + \mathbf{B}_k\mathbf{u}_k + \mathbf{w}_k
\end{equation}
\textbf{Measurement equation:}
\begin{equation}
\mathbf{z}_k = \mathbf{H}_k\mathbf{x}_k + \mathbf{v}_k
\end{equation}
Where:
\begin{itemize}
\item $\mathbf{x}_k$ is the state vector at time step $k$
\item $\mathbf{F}_k$ is the state transition matrix
\item $\mathbf{B}_k$ is the control input matrix
\item $\mathbf{u}_k$ is the control vector
\item $\mathbf{w}_k$ is the process noise, with covariance $\mathbf{Q}_k$
\item $\mathbf{z}_k$ is the measurement vector
\item $\mathbf{H}_k$ is the measurement matrix
\item $\mathbf{v}_k$ is the measurement noise, with covariance $\mathbf{R}_k$
\end{itemize}
The noise terms are assumed to be white, zero-mean, Gaussian, and independent of each other:
\begin{equation}
\mathbf{w}_k \sim \mathcal{N}(0, \mathbf{Q}_k)
\end{equation}
\begin{equation}
\mathbf{v}_k \sim \mathcal{N}(0, \mathbf{R}_k)
\end{equation}

\subsubsubsection{Probabilistic Interpretation}
The Kalman Filter maintains two probability distributions:
\begin{enumerate}
\item \textbf{Prior distribution:} $p(\mathbf{x}_k | \mathbf{z}_{1:k-1})$ - State estimate before incorporating measurement at time $k$
\item \textbf{Posterior distribution:} $p(\mathbf{x}_k | \mathbf{z}_{1:k})$ - State estimate after incorporating measurement at time $k$
\end{enumerate}
Both distributions are Gaussian and fully specified by their means and covariances:
\begin{itemize}
\item Prior: $\mathcal{N}(\hat{\mathbf{x}}_{k|k-1}, \mathbf{P}_{k|k-1})$
\item Posterior: $\mathcal{N}(\hat{\mathbf{x}}_{k|k}, \mathbf{P}_{k|k})$
\end{itemize}

\subsubsection{Kalman Filter Algorithm}
The Kalman Filter operates in a predict-update cycle:

\subsubsubsection{Prediction Step}
The prediction step projects the current state and error covariance forward in time to obtain the a priori estimates for the next time step: \\
\textbf{State prediction:}
\begin{equation}
\hat{\mathbf{x}}_{k|k-1} = \mathbf{F}_k\hat{\mathbf{x}}_{k-1|k-1} + \mathbf{B}_k\mathbf{u}_k
\end{equation}
\textbf{Covariance prediction:}
\begin{equation}
\mathbf{P}_{k|k-1} = \mathbf{F}_k\mathbf{P}_{k-1|k-1}\mathbf{F}_k^T + \mathbf{Q}_k
\end{equation}

\subsubsubsection{Update Step}
The update step incorporates a new measurement to obtain improved a posteriori estimates: \\
\textbf{Innovation (measurement residual):}
\begin{equation}
\mathbf{y}_k = \mathbf{z}_k - \mathbf{H}_k\hat{\mathbf{x}}_{k|k-1}
\end{equation}
\textbf{Innovation covariance:}
\begin{equation}
\mathbf{S}_k = \mathbf{H}_k\mathbf{P}_{k|k-1}\mathbf{H}_k^T + \mathbf{R}_k
\end{equation}
\textbf{Optimal Kalman gain:}
\begin{equation}
\mathbf{K}_k = \mathbf{P}_{k|k-1}\mathbf{H}_k^T\mathbf{S}_k^{-1}
\end{equation}
\textbf{Updated state estimate:}
\begin{equation}
\hat{\mathbf{x}}_{k|k} = \hat{\mathbf{x}}_{k|k-1} + \mathbf{K}_k\mathbf{y}_k
\end{equation}
\textbf{Updated covariance estimate:}
\begin{equation}
\mathbf{P}_{k|k} = (\mathbf{I} - \mathbf{K}_k\mathbf{H}_k)\mathbf{P}_{k|k-1}
\end{equation}
An alternative form for the covariance update that is numerically more stable is:
\begin{equation}
\mathbf{P}_{k|k} = (\mathbf{I} - \mathbf{K}_k\mathbf{H}_k)\mathbf{P}_{k|k-1}(\mathbf{I} - \mathbf{K}_k\mathbf{H}_k)^T + \mathbf{K}_k\mathbf{R}_k\mathbf{K}_k^T
\end{equation}

\begin{algorithm}
\caption{Kalman Filter}
\begin{algorithmic}[1]
\Require Initial state estimate $\hat{\mathbf{x}}_{0|0}$, initial error covariance $\mathbf{P}_{0|0}$,\
measurement sequence $\mathbf{z}_1, \mathbf{z}_2, \ldots, \mathbf{z}_n$,\
control inputs $\mathbf{u}_1, \mathbf{u}_2, \ldots, \mathbf{u}_n$ (if applicable),\
system matrices $\mathbf{F}, \mathbf{H}, \mathbf{B}, \mathbf{Q}, \mathbf{R}$
\Ensure Filtered state estimates $\hat{\mathbf{x}}_{1|1}, \hat{\mathbf{x}}_{2|2}, \ldots, \hat{\mathbf{x}}_{n|n}$ and\
their error covariances $\mathbf{P}_{1|1}, \mathbf{P}_{2|2}, \ldots, \mathbf{P}_{n|n}$
\State $\hat{\mathbf{x}}_{0|0} \gets$ initial state estimate
\State $\mathbf{P}_{0|0} \gets$ initial error covariance
\For{$k = 1, 2, \ldots, n$}
\State // Prediction Step
\State $\hat{\mathbf{x}}_{k|k-1} \gets \mathbf{F}_k\hat{\mathbf{x}}_{k-1|k-1} + \mathbf{B}_k\mathbf{u}_k$
\State $\mathbf{P}_{k|k-1} \gets \mathbf{F}_k\mathbf{P}_{k-1|k-1}\mathbf{F}_k^T + \mathbf{Q}_k$
\State // Update Step
\State $\mathbf{y}_k \gets \mathbf{z}_k - \mathbf{H}_k\hat{\mathbf{x}}_{k|k-1}$ \Comment{Measurement residual}
\State $\mathbf{S}_k \gets \mathbf{H}_k\mathbf{P}_{k|k-1}\mathbf{H}_k^T + \mathbf{R}_k$ \Comment{Residual covariance}
\State $\mathbf{K}_k \gets \mathbf{P}_{k|k-1}\mathbf{H}_k^T(\mathbf{S}_k)^{-1}$ \Comment{Optimal Kalman gain}
\State $\hat{\mathbf{x}}_{k|k} \gets \hat{\mathbf{x}}_{k|k-1} + \mathbf{K}_k\mathbf{y}_k$ \Comment{Updated state estimate}
\State $\mathbf{P}_{k|k} \gets (\mathbf{I} - \mathbf{K}_k\mathbf{H}_k)\mathbf{P}_{k|k-1}$ \Comment{Updated error covariance}

\State // Output filtered state estimate and covariance
\State \textbf{Output} $\hat{\mathbf{x}}_{k|k}, \mathbf{P}_{k|k}$
\EndFor
\end{algorithmic}
\end{algorithm}

\subsubsection{Optimal Properties of the Kalman Filter}
The Kalman Filter is optimal in the sense that:
\begin{enumerate}
\item If all noise is Gaussian, it minimizes the mean square error of the estimated parameters.
\item If the noise is not Gaussian but is uncorrelated, the Kalman Filter is the best linear unbiased estimator (BLUE).
\item The estimate is unbiased, meaning $E[\hat{\mathbf{x}}_k - \mathbf{x}_k] = 0$.
\item Among all possible filters, the Kalman Filter minimizes the trace of the error covariance matrix.
\end{enumerate}

\subsubsection{Interpretation of Kalman Gain}
The Kalman gain $\mathbf{K}_k$ determines how much the filter should "trust" the new measurement versus the predicted state. It can be interpreted as:
\begin{itemize}
\item When measurement noise is small ($\mathbf{R}_k \to 0$), the gain gives more weight to measurements: $\mathbf{K}_k \to \mathbf{H}_k^{-1}$
\item When process noise is small ($\mathbf{Q}_k \to 0$), the gain gives more weight to predictions: $\mathbf{K}_k \to 0$
\end{itemize}
This automatic weighting of predictions and measurements based on their respective uncertainties is one of the key strengths of the Kalman Filter.

\subsubsection{Application to Object Tracking}
For object tracking applications, the state vector typically includes position and velocity:
\begin{equation}
\mathbf{x} = [x, \dot{x}, y, \dot{y}]^T
\end{equation}
For a constant velocity model, the state transition matrix becomes:
\begin{equation}
\mathbf{F} = \begin{bmatrix}
    1 & \Delta t & 0 & 0 \\
    0 & 1 & 0 & 0 \\
    0 & 0 & 1 & \Delta t \\
    0 & 0 & 0 & 1
    \end{bmatrix}
\end{equation}
And if only position is measured, the measurement matrix is:
\begin{equation}
\mathbf{H} = \begin{bmatrix}
1 & 0 & 0 & 0 \\
0 & 0 & 1 & 0
\end{bmatrix}
\end{equation}
The process noise covariance $\mathbf{Q}$ accounts for model uncertainties, while the measurement noise covariance $\mathbf{R}$ represents sensor noise. These parameters need to be tuned based on the specific tracking scenario and sensor characteristics.

\section{Acknowledgment}
We gratefully acknowledge the Louisiana Transportation Research Center (LTRC) for funding the graduate research assistantship that made this work possible. We would also like to thank all the undergraduate research apprentices of the CPHS Lab at UL Lafayette for their contribution in data collection and labelling.   

Experiments were performed on UL Lafayette's VAStream Sandbox with Nvidia P100 GPUs. Any errors here are our own and don't reflect the opinions of our colleagues and proofreaders.
\vspace{0.5in}
\bibliographystyle{unsrt}  
\bibliography{references}  

\begin{thebibliography}{10}

\bibitem{fastrcnn}
Ross Girshick.
\newblock Fast r-cnn, 2015.

\bibitem{fasterrcnn}
Shaoqing Ren, Kaiming He, Ross Girshick, and Jian Sun.
\newblock Faster r-cnn: Towards real-time object detection with region proposal
  networks.
\newblock {\em IEEE transactions on pattern analysis and machine intelligence},
  39(6):1137--1149, 2016.

\bibitem{wang2023yolov7}
Chien-Yao Wang, Alexey Bochkovskiy, and Hong-Yuan~Mark Liao.
\newblock Yolov7: Trainable bag-of-freebies sets new state-of-the-art for
  real-time object detectors.
\newblock In {\em Proceedings of the IEEE/CVF conference on computer vision and
  pattern recognition}, pages 7464--7475, 2023.

\bibitem{yaseen2024yolov8indepthexplorationinternal}
Muhammad Yaseen.
\newblock What is yolov8: An in-depth exploration of the internal features of
  the next-generation object detector, 2024.

\bibitem{khanam2024yolov5deeplookinternal}
Rahima Khanam and Muhammad Hussain.
\newblock What is yolov5: A deep look into the internal features of the popular
  object detector, 2024.

\bibitem{ssd}
Wei Liu, Dragomir Anguelov, Dumitru Erhan, Christian Szegedy, Scott Reed,
  Cheng-Yang Fu, and Alexander~C. Berg.
\newblock {\em SSD: Single Shot MultiBox Detector}, page 21–37.
\newblock Springer International Publishing, 2016.

\bibitem{opticalflow}
Ryuzo Okada, Yoshiaki Shirai, and Jun Miura.
\newblock Object tracking based on optical flow and depth.
\newblock In {\em 1996 IEEE/SICE/RSJ International Conference on Multisensor
  Fusion and Integration for Intelligent Systems (Cat. No. 96TH8242)}, pages
  565--571. IEEE, 1996.

\bibitem{meanshift}
Dorin Comaniciu, Visvanathan Ramesh, and Peter Meer.
\newblock Real-time tracking of non-rigid objects using mean shift.
\newblock In {\em Proceedings IEEE Conference on Computer Vision and Pattern
  Recognition. CVPR 2000 (Cat. No. PR00662)}, volume~2, pages 142--149. IEEE,
  2000.

\bibitem{correlationfilter}
Jack Valmadre, Luca Bertinetto, Joao Henriques, Andrea Vedaldi, and Philip~HS
  Torr.
\newblock End-to-end representation learning for correlation filter based
  tracking.
\newblock In {\em Proceedings of the IEEE conference on computer vision and
  pattern recognition}, pages 2805--2813, 2017.

\bibitem{bewley2016simple}
Alex Bewley, Zongyuan Ge, Lionel Ott, Fabio Ramos, and Ben Upcroft.
\newblock Simple online and realtime tracking.
\newblock In {\em 2016 IEEE international conference on image processing
  (ICIP)}, pages 3464--3468. IEEE, 2016.

\bibitem{maggiolino2023deep}
Gerard Maggiolino, Adnan Ahmad, Jinkun Cao, and Kris Kitani.
\newblock Deep oc-sort: Multi-pedestrian tracking by adaptive
  re-identification.
\newblock {\em arXiv preprint arXiv:2302.11813}, 2023.

\bibitem{cao2023observation}
Jinkun Cao, Jiangmiao Pang, Xinshuo Weng, Rawal Khirodkar, and Kris Kitani.
\newblock Observation-centric sort: Rethinking sort for robust multi-object
  tracking.
\newblock In {\em Proceedings of the IEEE/CVF Conference on Computer Vision and
  Pattern Recognition}, pages 9686--9696, 2023.

\bibitem{zhang2022bytetrack}
Yifu Zhang, Peize Sun, Yi~Jiang, Dongdong Yu, Fucheng Weng, Zehuan Yuan, Ping
  Luo, Wenyu Liu, and Xinggang Wang.
\newblock Bytetrack: Multi-object tracking by associating every detection box.
\newblock In {\em Computer Vision--ECCV 2022: 17th European Conference, Tel
  Aviv, Israel, October 23--27, 2022, Proceedings, Part XXII}, pages 1--21.
  Springer, 2022.

\bibitem{aharon2022bot}
Nir Aharon, Roy Orfaig, and Ben-Zion Bobrovsky.
\newblock Bot-sort: Robust associations multi-pedestrian tracking.
\newblock {\em arXiv preprint arXiv:2206.14651}, 2022.

\bibitem{du2023strongsort}
Yunhao Du, Zhicheng Zhao, Yang Song, Yanyun Zhao, Fei Su, Tao Gong, and
  Hongying Meng.
\newblock Strongsort: Make deepsort great again.
\newblock {\em IEEE Transactions on Multimedia}, 2023.

\bibitem{GenTracker}
Michael~J Black and Allan~D Jepson.
\newblock Eigentracking: Robust matching and tracking of articulated objects
  using a view-based representation.
\newblock {\em International Journal of Computer Vision}, 26:63--84, 1998.

\bibitem{Discriminative1}
Boris Babenko, Ming-Hsuan Yang, and Serge Belongie.
\newblock Visual tracking with online multiple instance learning.
\newblock In {\em 2009 IEEE Conference on computer vision and Pattern
  Recognition}, pages 983--990. IEEE, 2009.

\bibitem{Discriminative2}
Shai Avidan.
\newblock Support vector tracking.
\newblock {\em IEEE transactions on pattern analysis and machine intelligence},
  26(8):1064--1072, 2004.

\bibitem{Discriminative3}
Hyeonseob Nam and Bohyung Han.
\newblock Learning multi-domain convolutional neural networks for visual
  tracking.
\newblock In {\em Proceedings of the IEEE conference on computer vision and
  pattern recognition}, pages 4293--4302, 2016.

\bibitem{Discriminative4}
Jo{\~a}o~F Henriques, Rui Caseiro, Pedro Martins, and Jorge Batista.
\newblock High-speed tracking with kernelized correlation filters.
\newblock {\em IEEE transactions on pattern analysis and machine intelligence},
  37(3):583--596, 2014.

\bibitem{collaborative1}
Helmut Grabner, Christian Leistner, and Horst Bischof.
\newblock Semi-supervised on-line boosting for robust tracking.
\newblock In {\em Computer Vision--ECCV 2008: 10th European Conference on
  Computer Vision, Marseille, France, October 12-18, 2008, Proceedings, Part I
  10}, pages 234--247. Springer, 2008.

\bibitem{collaborative2}
Xue Mei, Haibin Ling, Yi~Wu, Erik Blasch, and Li~Bai.
\newblock Minimum error bounded efficient l1 tracker with occlusion detection.
\newblock In {\em CVPR 2011}, pages 1257--1264. IEEE, 2011.

\bibitem{surveySOD}
Gong Cheng, Xiang Yuan, Xiwen Yao, Kebing Yan, Qinghua Zeng, Xingxing Xie, and
  Junwei Han.
\newblock Towards large-scale small object detection: Survey and benchmarks.
\newblock {\em IEEE Transactions on Pattern Analysis and Machine Intelligence},
  2023.

\bibitem{visDrone}
Pengfei Zhu, Longyin Wen, Dawei Du, Xiao Bian, Heng Fan, Qinghua Hu, and Haibin
  Ling.
\newblock Detection and tracking meet drones challenge.
\newblock {\em IEEE Transactions on Pattern Analysis and Machine Intelligence},
  44(11):7380--7399, 2021.

\bibitem{DOTA}
Jian Ding, Nan Xue, Gui-Song Xia, Xiang Bai, Wen Yang, Michael~Ying Yang, Serge
  Belongie, Jiebo Luo, Mihai Datcu, Marcello Pelillo, et~al.
\newblock Object detection in aerial images: A large-scale benchmark and
  challenges.
\newblock {\em IEEE transactions on pattern analysis and machine intelligence},
  44(11):7778--7796, 2021.

\bibitem{tracknet}
Yu-Chuan Huang, I-No Liao, Ching-Hsuan Chen, Ts{\`\i}-U{\'\i} {\.I}k, and
  Wen-Chih Peng.
\newblock Tracknet: A deep learning network for tracking high-speed and tiny
  objects in sports applications.
\newblock In {\em 2019 16th IEEE International Conference on Advanced Video and
  Signal Based Surveillance (AVSS)}, pages 1--8. IEEE, 2019.

\bibitem{sun2020tracknetv2}
Nien-En Sun, Yu-Ching Lin, Shao-Ping Chuang, Tzu-Han Hsu, Dung-Ru Yu, Ho-Yi
  Chung, and Ts{\`\i}-U{\'\i} {\.I}k.
\newblock Tracknetv2: Efficient shuttlecock tracking network.
\newblock In {\em 2020 International Conference on Pervasive Artificial
  Intelligence (ICPAI)}, pages 86--91. IEEE, 2020.

\bibitem{ju2020trajectory}
Nyan-Ping Ju, Dung-Ru Yu, Tsi-Ui Ik, and Wen-Chih Peng.
\newblock Trajectory-based badminton shots detection.
\newblock In {\em 2020 International Conference on Pervasive Artificial
  Intelligence (ICPAI)}, pages 64--71. IEEE, 2020.

\bibitem{golfball}
Xiaohan Zhang, Tianxiao Zhang, Yiju Yang, Zongbo Wang, and Guanghui Wang.
\newblock Real-time golf ball detection and tracking based on convolutional
  neural networks.
\newblock In {\em 2020 IEEE International Conference on Systems, Man, and
  Cybernetics (SMC)}, pages 2808--2813. IEEE, 2020.

\bibitem{kuhn1955hungarian}
Harold~W Kuhn.
\newblock The hungarian method for the assignment problem.
\newblock {\em Naval research logistics quarterly}, 2(1-2):83--97, 1955.

\bibitem{bernardin2008evaluating}
Keni Bernardin and Rainer Stiefelhagen.
\newblock Evaluating multiple object tracking performance: the clear mot
  metrics.
\newblock {\em EURASIP Journal on Image and Video Processing}, 2008:1--10,
  2008.

\bibitem{ristani2016performance}
Ergys Ristani, Francesco Solera, Roger Zou, Rita Cucchiara, and Carlo Tomasi.
\newblock Performance measures and a data set for multi-target, multi-camera
  tracking.
\newblock In {\em European conference on computer vision}, pages 17--35.
  Springer, 2016.

\bibitem{luiten2021hota}
Jonathon Luiten, Aljosa Osep, Patrick Dendorfer, Philip Torr, Andreas Geiger,
  Laura Leal-Taix{\'e}, and Bastian Leibe.
\newblock Hota: A higher order metric for evaluating multi-object tracking.
\newblock {\em International journal of computer vision}, 129:548--578, 2021.

\bibitem{li2009learning}
Yuan Li, Chang Huang, and Ram Nevatia.
\newblock Learning to associate: Hybridboosted multi-target tracker for crowded
  scene.
\newblock In {\em 2009 IEEE conference on computer vision and pattern
  recognition}, pages 2953--2960. IEEE, 2009.

\bibitem{ADE_metrics}
Dmytro Zabolotnii, Yar Muhammad, and Naveed Muhammad.
\newblock Pedestrian motion prediction evaluation for urban autonomous driving,
  2024.

\bibitem{mohamed2022social}
Abduallah Mohamed, Deyao Zhu, Warren Vu, Mohamed Elhoseiny, and Christian
  Claudel.
\newblock Social-implicit: Rethinking trajectory prediction evaluation and the
  effectiveness of implicit maximum likelihood estimation.
\newblock In {\em European Conference on Computer Vision}, pages 463--479.
  Springer, 2022.

\bibitem{SORT_Pic}
Ricardo Pereira, Guilherme Carvalho, Luís Garrote, and Urbano Nunes.
\newblock Sort and deep-sort based multi-object tracking for mobile robotics:
  Evaluation with new data association metrics.
\newblock {\em Applied Sciences}, 12:1319, 01 2022.

\bibitem{github}
Ifzhang.
\newblock Bytetrack: Multi-object tracking by associating every detection box,
  2020.

\bibitem{fischler1981random}
MA~FISCHLER~AND.
\newblock Random sample consensus: a paradigm for model fitting with
  applications to image analysis and automated cartography.
\newblock {\em Commun. ACM}, 24(6):381--395, 1981.

\bibitem{wojke2017simple}
Nicolai Wojke, Alex Bewley, and Dietrich Paulus.
\newblock Simple online and realtime tracking with a deep association metric.
\newblock In {\em 2017 IEEE international conference on image processing
  (ICIP)}, pages 3645--3649. IEEE, 2017.

\bibitem{kalman1960new}
Rudolph~Emil Kalman.
\newblock A new approach to linear filtering and prediction problems.
\newblock {\em Journal of Basic Engineering.}, 1960.

\bibitem{thrun2002probabilistic}
Sebastian Thrun.
\newblock Probabilistic robotics.
\newblock {\em Communications of the ACM}, 45(3):52--57, 2002.

\end{thebibliography}

\end{document}